%% file: main.tex
\documentclass[preprint,12pt]{elsarticle}

\usepackage{amsmath}
\usepackage{amssymb}
\usepackage{graphicx}
\usepackage{booktabs}
\usepackage[section]{placeins}
\usepackage{flafter}
\usepackage{float}

\begin{document}

\begin{frontmatter}
\title{A Computational Comparison of Fourier Spectral Differentiation and Spatial Automatic Differentiation in Periodic Physics-Informed Neural Networks}
\author[1]{Xilai Liang\corref{cor1}}
\ead{liang24532@gtiit.edu.cn}
\author[2]{Zhao Zhang\corref{cor1}}
\ead{zhaozhang@sdu.edu.cn}
\cortext[cor1]{Corresponding authors}
\address[1]{Guangdong Technion--Israel Institute of Technology,
Shantou, Guangdong, China}
\address[2]{Research Centre for Mathematics and Interdisciplinary Sciences, Shandong University, Qingdao, Shandong Province, 266237, China}

\begin{abstract}
Physics-informed neural networks (PINNs) commonly evaluate the spatial derivatives appearing in partial differential equation residuals using automatic differentiation (AD), whose computational and memory costs can become substantial when multiple or high-order derivatives are required. We perform a controlled comparison of spatial AD and Fourier spectral differentiation in periodic physical-space PINNs. Within each paired experiment, the neural representation, temporal differentiation, optimizer, sampling procedure, and training schedule are held fixed, so that the two cases differ only in the spatial differentiation procedure. For the Fourier variant, network outputs are evaluated on a uniform periodic grid and transformed to Fourier space, where spatial derivatives are obtained through spectral multiplication and the same Fourier coefficients are reused across derivative orders. We compare the two procedures in standard PINNs for the Allen--Cahn and Korteweg--de Vries equations and in Causal PINNs for the Allen--Cahn, Korteweg--de Vries, and Kuramoto--Sivashinsky equations. Across these five equation--framework settings, Fourier differentiation
yields mean paired end-to-end training speedups ranging from
\(2.90\times\) to \(18.52\times\) and reduces peak allocated graphics
processing unit (GPU) memory by \(68.7\%\)--\(94.1\%\). The final relative $L_2$ errors remain of the same order, with neither differentiation procedure showing a consistent accuracy advantage. For the one-dimensional periodic benchmarks considered here, Fourier spectral differentiation therefore provides substantially lower training time and memory usage than spatial AD while retaining comparable solution error, at the cost of requiring a uniform structured spatial grid.
\end{abstract}
\end{frontmatter}
\input{sections/01_introduction}
\input{sections/02_method}
\input{sections/03_experiments}
\input{sections/04_results}
\input{sections/05_discussion}
\input{sections/06_conclusion}

\bibliographystyle{unsrt}
\bibliography{references}

\end{document}

%% file: sections/01_introduction.tex
\section{Introduction}

Physics-informed neural networks (PINNs) approximate solutions to partial differential equations (PDEs) by representing the unknown field with a neural network and incorporating the governing equations, initial conditions, and boundary conditions into the training objective~\cite{raissi2019pinn}. In coordinate-based PINNs, the derivatives required by the PDE residual are commonly evaluated by automatic differentiation (AD) with respect to the network inputs~\cite{raissi2019pinn}. This provides pointwise derivatives without introducing an external spatial discretization, but derivative evaluation is required at every residual computation. When a PDE contains multiple spatial derivative terms or high-order derivatives, repeated AD introduces additional differentiation operations and larger derivative graphs, increasing both computational and memory costs during training.

Several approaches reduce or avoid this dependence on spatial AD through numerical differentiation or reformulation of the governing equations. Sharma and Shankar~\cite{sharma2022dtpinn} used radial basis function finite-difference discretizations to evaluate spatial derivatives while retaining AD for temporal derivatives in time-dependent problems. The coupled automatic--numerical differentiation physics-informed neural network (CAN-PINN) combines AD and numerical differentiation to couple neighboring support points~\cite{chiu2022canpinn}, whereas the smoothing-kernel physics-informed neural network (SK-PINN) evaluates derivatives through smoothing-kernel discretization~\cite{pan2025skpinn}. A different strategy is to reduce repeated high-order differentiation by rewriting higher-order PDEs as first-order systems, as in first-order physics-informed neural networks~\cite{gladstone2025fopinn}. More recently, FlashPDE has
treated derivative evaluation as a differentiable grid-based operator layer
and implemented fused finite-difference operators independently of the
surrounding neural architecture~\cite{zang2026flashpde}.

Spectral discretizations provide another route for evaluating the derivatives used in physics-informed objectives. Pseudo-spectral PINN formulations have employed spectral discretization for
physics-informed model discovery~\cite{zhao2021pseudospectral}, while neural
spectral element methods evaluate neural fields on fixed spectral nodes and
replace derivative calls with spectral differentiation
matrices~\cite{feugmo2026nsem}. The Spectral Informed Neural Network (SINN) represents the solution through Fourier coefficients and converts spatial differentiation into multiplication in the spectral domain~\cite{yu2026sinn}. Other recent approaches, such as trainable Fourier feature-grid formulations, also use Fourier-transform-based derivative evaluation within modified neural representations~\cite{zhao2026beignet}. Xiao et al.~\cite{xiao2022fourier} replaced the finite-difference filter in a physics-informed convolutional recurrent network (PhyCRNet) with a Fourier filter, transforming network outputs to Fourier space for spatial differentiation and returning the resulting quantities through the inverse transform for construction of the physics-informed loss. These studies establish that numerical and spectral differentiation can be incorporated into physics-informed learning through several neural representations and operator constructions.

This study addresses a controlled computational question: when the surrounding physical-space PINN formulation is held fixed, how much of the lower derivative cost of Fourier spectral differentiation translates into end-to-end training-time and memory savings relative to spatial AD, and how does the change affect solution accuracy? A cheaper derivative operator does not necessarily yield a proportional end-to-end speedup when network evaluation or other components dominate the training loop; Fourier filter-based PhyCRNet reports such behavior when its time-series module accounts for most of the computational cost~\cite{xiao2022fourier}. We examine this question in both standard PINNs and Causal PINNs, where the latter use the temporal causal weighting formulation of Wang et al.~\cite{wang2024causal}. Standard PINNs are evaluated on the Allen--Cahn and Korteweg--de Vries (KdV) equations, and Causal PINNs are evaluated on Allen--Cahn, KdV, and Kuramoto--Sivashinsky (KS). Each equation--framework setting uses three paired random seeds, with all components of each paired training setup held fixed except the spatial differentiation procedure. We compare final relative $L_2$ error, end-to-end training time, and peak allocated memory. Across the five equation--framework settings, Fourier spectral differentiation yields mean paired speedups ranging from $2.90\times$ to $18.52\times$ and reduces peak allocated memory by $68.7\%$--$94.1\%$. The final relative $L_2$ errors remain of the same order, with neither differentiation procedure showing a consistent accuracy advantage. These results characterize the computational trade-off between spatial AD and Fourier spectral differentiation for the one-dimensional periodic PINNs considered here, where the Fourier procedure requires a uniform structured spatial grid.

%% file: sections/02_method.tex
\section{Spatial Differentiation Procedures for Physical-Space PINNs}

We consider a time-dependent PDE of the form
\begin{equation}
u_t + \mathcal{N}(u,u_x,u_{xx},\ldots)=0,
\end{equation}
where the solution is represented by a physical-space neural network
\(u_\theta=u_\theta(x,t)\).
We compare two procedures for evaluating the spatial derivatives entering
the PDE residual: automatic differentiation and Fourier spectral
differentiation. The neural representation and temporal differentiation
procedure are identical in the two cases.

\subsection{Standard and Causal PINN Formulations}

For a sampled time location \(t_i\), the spatially averaged residual loss is
\begin{equation}
\mathcal{L}^{(i)}_t
=
\frac{1}{N_x}
\sum_{j=1}^{N_x}
r_\theta(x_j,t_i)^2,
\end{equation}
where \(r_\theta\) denotes the PDE residual. The initial-condition
contribution is evaluated on the same spatial grid as
\[
\mathcal{L}_0
=
10^4
\frac{1}{N_x}
\sum_{j=1}^{N_x}
\left[
u_\theta(x_j,0)-u_0(x_j)
\right]^2 .
\]

For the standard PINN, the temporal residual contributions are weighted
uniformly,
\begin{equation}
\mathcal{L}_{\mathrm{standard}}
=
\mathcal{L}_0
+
\frac{1}{N_t}
\sum_{i=1}^{N_t}
\mathcal{L}^{(i)}_t .
\end{equation}

For the Causal PINN, we follow the temporal causal-weighting construction
of Wang et al.~\cite{wang2024causal}. Let
\[
\mathbf{L}_t
=
\begin{bmatrix}
\mathcal{L}^{(1)}_t &
\mathcal{L}^{(2)}_t &
\cdots &
\mathcal{L}^{(N_t)}_t
\end{bmatrix}^{T},
\]
and let \(M\) denote the strictly lower-triangular accumulation matrix,
with \(M_{ij}=1\) for \(j<i\) and \(M_{ij}=0\) otherwise. The causal
weights are evaluated as
\begin{equation}
\mathbf{w}
=
\exp\left[
-\epsilon
\left(
M\mathbf{L}_t
+
\mathcal{L}_0\mathbf{1}
\right)
\right],
\end{equation}
where \(\mathbf{1}\in\mathbb{R}^{N_t}\) denotes the vector of ones and the
exponential is applied elementwise. We use a fixed causality parameter
\(\epsilon=0.1\) throughout training. The causal weights are detached from
the computational graph before evaluating the weighted residual loss, giving
\begin{equation}
\mathcal{L}_{\mathrm{causal}}
=
\mathcal{L}_0
+
\frac{1}{N_t}
\sum_{i=1}^{N_t}
w_i\mathcal{L}^{(i)}_t .
\end{equation}

\subsection{Spatial Automatic Differentiation}

In the automatic-differentiation baseline, spatial derivatives are
obtained by repeated differentiation of the physical-space network output.
For example,
\[
u_x
=
\frac{\partial u_\theta}{\partial x},
\qquad
u_{xx}
=
\frac{\partial}{\partial x}
\left(
\frac{\partial u_\theta}{\partial x}
\right),
\]
with higher-order derivatives obtained through additional nested
differentiation operations. Only the spatial derivative orders appearing
in the corresponding PDE residual are evaluated. In our implementation,
these derivatives are constructed using nested Jacobian--vector products
(JVPs). Temporal differentiation is independent of the spatial
differentiation procedure, and \(u_t\) is evaluated through the same
JVP-based automatic-differentiation path in both cases.

\subsection{Fourier Spectral Differentiation}

For Fourier spectral differentiation, the network is evaluated on an
endpoint-excluded uniform periodic grid,
\[
x_j=x_0+j\Delta x,
\qquad
j=0,\ldots,N_x-1,
\qquad
\Delta x=\frac{L}{N_x},
\]
where \(L\) is the spatial period. At a fixed time \(t\), the sampled
network output is transformed to Fourier space,
\[
\widehat{u}_k(t)
=
\mathcal{F}_x
\left[
u_\theta(x_j,t)
\right],
\]
where \(\mathcal{F}_x\) denotes the discrete Fourier transform in the
spatial direction. For the one-sided real Fourier representation used here, the discrete
wavenumbers are
\[
k_n
=
\frac{2\pi n}{L},
\qquad
n=0,\ldots,\frac{N_x}{2},
\]
for the even spatial resolution used in the experiments.
An \(m\)th-order spatial derivative is then evaluated as
\begin{equation}
\partial_x^m u
=
\mathcal{F}_x^{-1}
\left[
(ik)^m\widehat{u}_k
\right].
\end{equation}
For even \(N_x\), the real inverse transform enforces the Hermitian
constraint at the Nyquist mode. Consequently, the Nyquist contribution
vanishes for odd-order spatial derivatives and is retained for even-order
derivatives.

At finite \(N_x\), this operation differentiates the periodic
trigonometric interpolant defined by the sampled network values. When
multiple spatial derivative orders are required at the same time
location, a single Fourier transform of the network output is reused and
each derivative is obtained from the same Fourier coefficients using the
corresponding spectral multiplier. The derivatives are then transformed
back to physical space and used to construct the PDE residual.

The transforms are implemented using the real-valued fast Fourier
transform routines provided by PyTorch. These operations remain within
the computational graph, so gradients of the physics-informed loss
propagate through the spectral differentiation procedure to the network
parameters.

%% file: sections/03_experiments.tex
\section{Numerical Experiments}

\subsection{Benchmark Problems}

We consider three one-dimensional periodic PDEs: the Allen--Cahn, KdV, and
KS equations. These equations contain different combinations and orders of
spatial derivatives and are used to compare automatic differentiation and
Fourier spectral differentiation under the two PINN formulations defined
in Section~2.

The Allen--Cahn equation is
\begin{equation}
u_t
-
10^{-4}u_{xx}
+
5u^3
-
5u
=
0,
\end{equation}
on \(x\in[-1,1)\) and \(t\in[0,0.1]\), with initial condition
\(u(x,0)=0.5\cos(\pi x)\). The KdV equation is
\begin{equation}
u_t
+
u u_x
+
0.0025u_{xxx}
=
0,
\end{equation}
on \(x\in[-1,1)\) and \(t\in[0,0.1]\), with initial condition
\(u(x,0)=\cos(\pi x)\). The KS equation used in the experiments is
\begin{equation}
u_t
+
\frac{100}{16}u u_x
+
\frac{100}{16^2}u_{xx}
+
\frac{100}{16^4}u_{xxxx}
=
0,
\end{equation}
on \(x\in[0,2\pi)\) and \(t\in[0,0.1]\). Its initial condition is taken
from the first time slice of the frozen KS numerical trajectory distributed
with the accompanying Causal PINN code and data of Wang et al.~\cite{wang2024causal}.
The spatial derivative sets required by Allen--Cahn, KdV, and KS are
\(u_{xx}\), \((u_x,u_{xxx})\), and
\((u_x,u_{xx},u_{xxxx})\), respectively.

\subsection{Experimental Setup}

All experiments use the same underlying physical-space neural-network
configuration. Each spatial domain uses \(N_x=256\) uniform periodic grid
points. The neural representation is an eight-layer gated multilayer
perceptron with width 128, tanh gating, Fourier input encodings with
\(m_t=6\) and \(m_x=5\), and a scalar output. The spatial coordinate enters
the network only through periodic sine--cosine Fourier features with the
period of the corresponding spatial domain. This representation enforces periodicity in the spatial coordinate, so no
separate spatial boundary-condition loss is used. The model contains 122,625
trainable parameters. At each optimization step, \(N_t=32\) independent
temporal samples are drawn uniformly from the training interval. For each
equation and seed, the automatic-differentiation and Fourier runs use the
same pre-generated temporal sampling schedule. The temporal samples are
sorted before causal weighting in the Causal PINN experiments.

Training uses the Adam optimizer~\cite{kingma2015adam}. We use
\(\eta_s = 10^{-3}\,0.9^{s/5000}\), where \(s\) denotes the optimization
step. The Adam parameters are \(\beta_1=0.9\), \(\beta_2=0.999\), and
\(\epsilon_{\mathrm{Adam}}=10^{-8}\), with zero weight decay and no gradient
clipping. All runs use single-precision floating-point arithmetic and are
trained for 50,000 optimization steps without early stopping. Three random
seeds, 1234, 2345, and 3456, are used for each equation--framework setting.
The standard PINN experiments are performed for Allen--Cahn and KdV using
the three paired seeds, giving 12 runs in total. The Causal PINN experiments
use the fixed causality parameter \(\epsilon=0.1\) defined in Section~2.1
and are performed for Allen--Cahn, KdV, and KS using the same three paired
seeds, giving 18 runs in total.

Within each paired comparison, the network initialization, temporal
sampling schedule, neural-network configuration, optimizer settings,
reference data, and training length are matched; the spatial differentiation
procedure is the only controlled change. The Causal PINN pairs also use the
same causal weighting configuration. The experiments were executed on an
NVIDIA GeForce RTX 4060 Laptop graphics processing unit (GPU) with 8 GB of
device memory using PyTorch 2.9.0 and CUDA 13.0.

\subsection{Reference Solutions and Evaluation}

Reference solutions for Allen--Cahn and KdV are generated independently
using a Fourier pseudospectral exponential time-differencing fourth-order
Runge--Kutta solver~\cite{kassam2005fourth} and are used only for
evaluation. The reference
calculations use an endpoint-excluded periodic grid with
\(N_x^{\mathrm{ref}}=2048\) and a time step
\(\Delta t=5\times10^{-5}\). Solutions are stored every \(0.001\) time
units, yielding 101 snapshots over \(t\in[0,0.1]\). The calculations use
double-precision real and complex arithmetic, and nonlinear terms are
dealiased using the two-thirds rule. The reference configurations were
checked through spatial and temporal refinement tests before evaluation.
For KS, the frozen numerical trajectory accompanying the Causal PINN code
and data of Wang et al.~\cite{wang2024causal} is used as the reference solution. The original
spatial data contain 512 points including a repeated periodic endpoint.
The repeated endpoint is removed and the solution is periodically
interpolated onto the \(N_x=256\) evaluation grid, giving 26 reference time
slices over \(t\in[0,0.1]\).

Solution accuracy is measured by the relative \(L_2\) error
\begin{equation}
E_{L_2}
=
\frac{
\left\|
u_\theta-u_{\mathrm{ref}}
\right\|_2
}{
\left\|
u_{\mathrm{ref}}
\right\|_2
}.
\end{equation}
For KS, this quantity is evaluated over the complete
\(26\times256\) space--time evaluation array; the reported final relative
\(L_2\) error therefore refers to the final trained model evaluated over the
fixed space--time reference set rather than to the error at the final time
slice alone.

Training time is measured over the complete 50,000-step training loop using
\texttt{time.perf\_counter()}. CUDA execution is synchronized around each
optimization step and once again at the end of training. The recorded time
includes scheduled model evaluation and standard-output logging inside the
training loop, while model initialization, warm-up, and post-training
artifact generation are excluded. For each paired comparison, the training
speedup is defined as
\begin{equation}
S
=
\frac{T_{\mathrm{AD}}}{T_{\mathrm{Fourier}}}.
\end{equation}

Peak allocated GPU memory is measured after resetting the CUDA peak-memory
statistics immediately before the training loop. Peak allocated memory is
reported in gibibytes (GiB). The memory reduction relative to automatic
differentiation is defined as
\begin{equation}
R_M
=
1-
\frac{M_{\mathrm{Fourier}}}{M_{\mathrm{AD}}}.
\end{equation}

%% file: sections/04_results.tex
\section{Results}
\label{sec:results}

We report the paired AD--Fourier comparisons first for standard PINNs and
then for Causal PINNs. Each equation--framework setting contains three
paired random seeds. Aggregate values are reported as the mean
\(\pm\) sample standard deviation (SD) over the three seeds. Seed 1234 is
used for the solution-field visualizations. Table~\ref{tab:results_summary}
summarizes the final solution error, paired training speedup, and
peak-memory reduction across all five equation--framework settings.

\begin{table}[t]
    \centering
    \caption{Summary of the paired AD--Fourier comparisons. Final relative
    \(L_2\) errors and training speedups are reported as the mean
    \(\pm\) sample standard deviation over three seeds. Memory reduction is
    based on peak allocated GPU memory.}
    \label{tab:results_summary}

    \small
    \setlength{\tabcolsep}{4.5pt}
    \renewcommand{\arraystretch}{1.08}

    \begin{tabular}{@{}lcccc@{}}
        \hline
        PDE
        & \multicolumn{2}{c}{Final relative \(L_2\) error \((\times10^{-3})\)}
        & Speedup
        & Memory \\
        \cline{2-3}
        &
        AD
        & Fourier
        & \(T_{\mathrm{AD}}/T_{\mathrm{Fourier}}\)
        & Reduction (\%) \\
        \hline

        \multicolumn{5}{@{}l}{\textit{Standard PINN}} \\[1pt]

        Allen--Cahn
        & \(1.322\pm0.565\)
        & \(1.238\pm0.653\)
        & \(3.19\pm0.28\)
        & \(68.7\) \\

        KdV
        & \(0.480\pm0.312\)
        & \(0.713\pm0.389\)
        & \(7.85\pm0.21\)
        & \(86.2\) \\[2pt]

        \multicolumn{5}{@{}l}{\textit{Causal PINN}} \\[1pt]

        Allen--Cahn
        & \(1.013\pm0.732\)
        & \(1.210\pm0.329\)
        & \(2.90\pm0.09\)
        & \(68.7\) \\

        KdV
        & \(1.366\pm0.905\)
        & \(0.514\pm0.222\)
        & \(7.13\pm0.16\)
        & \(86.2\) \\

        KS
        & \(4.404\pm4.056\)
        & \(2.870\pm0.920\)
        & \(18.52\pm2.65\)
        & \(94.1\) \\

        \hline
    \end{tabular}
\end{table}

\subsection{Standard PINN Results}

\begin{figure}[!htbp]
    \centering
    \includegraphics[width=\linewidth]
    {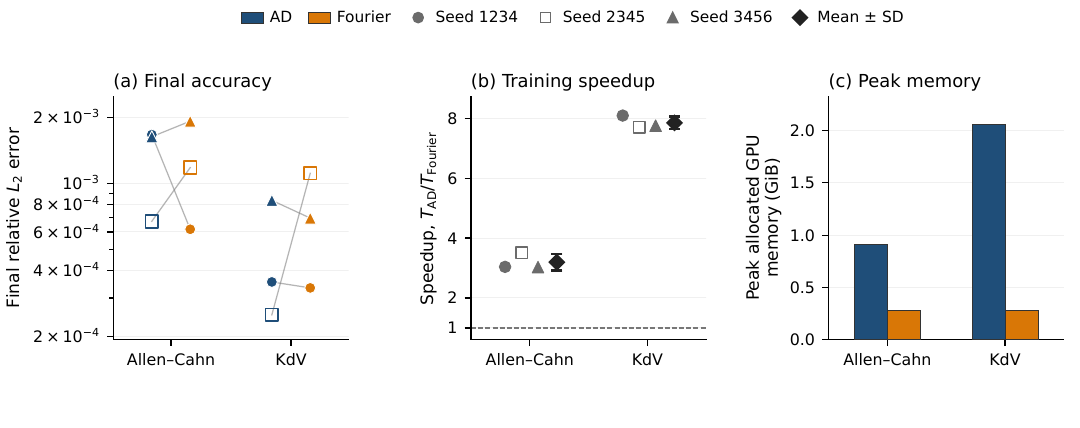}
    \caption{Quantitative comparison for the standard PINN experiments.
    (a) Final-model relative \(L_2\) errors for the three paired random
    seeds, with AD and Fourier results for the same seed connected within
    each equation. (b) Paired training speedup
    \(T_{\mathrm{AD}}/T_{\mathrm{Fourier}}\); individual markers denote
    the three seeds, while the diamond marker and error bar show the mean
    and sample standard deviation. The dashed line denotes a speedup of
    one. (c) Peak allocated GPU memory for AD and Fourier spectral
    differentiation; the peak-memory values are identical across the three
    seeds for each equation--method combination.}
    \label{fig:standard_summary}
\end{figure}

Figure~\ref{fig:standard_summary}(a) shows the paired final-model relative
\(L_2\) errors for the standard PINN experiments. Both differentiation
procedures reach errors of the same order for Allen--Cahn and KdV, with
visible variation across random seeds. Across the six paired comparisons,
AD gives the lower final error in three runs and Fourier differentiation
gives the lower final error in three runs. The aggregate values are reported
in Table~\ref{tab:results_summary}; neither differentiation procedure shows
a consistent accuracy advantage in the standard PINN experiments.

Figure~\ref{fig:standard_summary}(b,c) summarizes the computational
comparison. Fourier spectral differentiation reduces end-to-end training
time for every paired seed, with mean speedups of
\(3.19\pm0.28\times\) for Allen--Cahn and
\(7.85\pm0.21\times\) for KdV. Peak allocated GPU memory decreases from
approximately \(0.908\) to \(0.285\) GiB for Allen--Cahn and from
\(2.059\) to \(0.285\) GiB for KdV, corresponding to reductions of
\(68.7\%\) and \(86.2\%\), respectively.

\begin{figure}[!htbp]
    \centering
    \includegraphics[width=\linewidth]
    {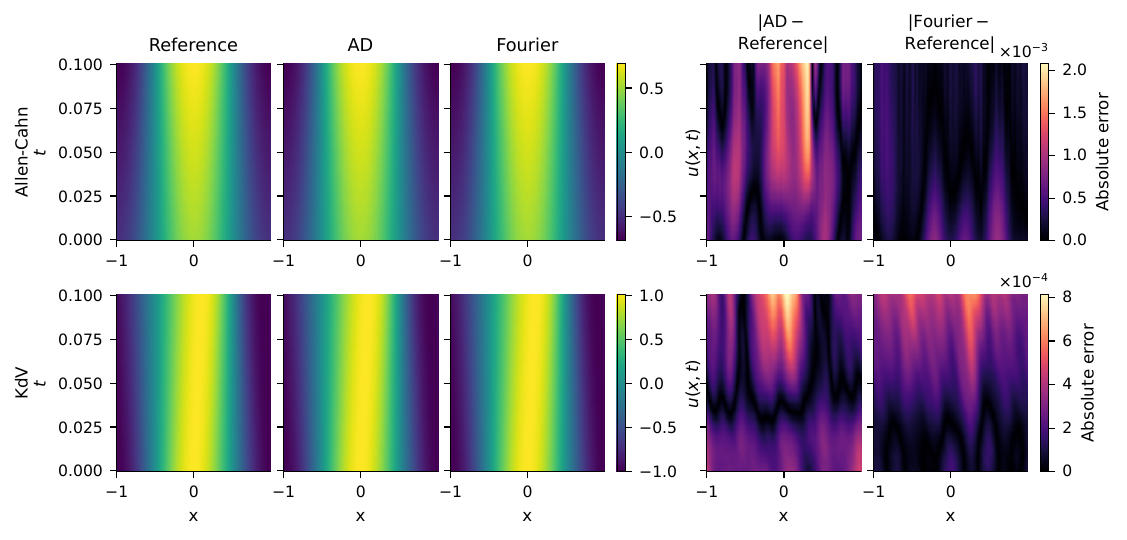}
    \caption{Reference, AD, and Fourier standard-PINN solutions and
    pointwise absolute errors for seed 1234. For each equation, the three
    solution panels share one color scale and the two error panels share
    another.}
    \label{fig:standard_fields}
\end{figure}

Figure~\ref{fig:standard_fields} compares the reference solutions with the
final predictions for seed 1234. Both differentiation procedures reproduce
the reference space--time structures for the displayed Allen--Cahn and KdV
solutions.

\FloatBarrier

\subsection{Causal PINN Results}

\begin{figure}[!htbp]
    \centering
    \includegraphics[width=\linewidth]
    {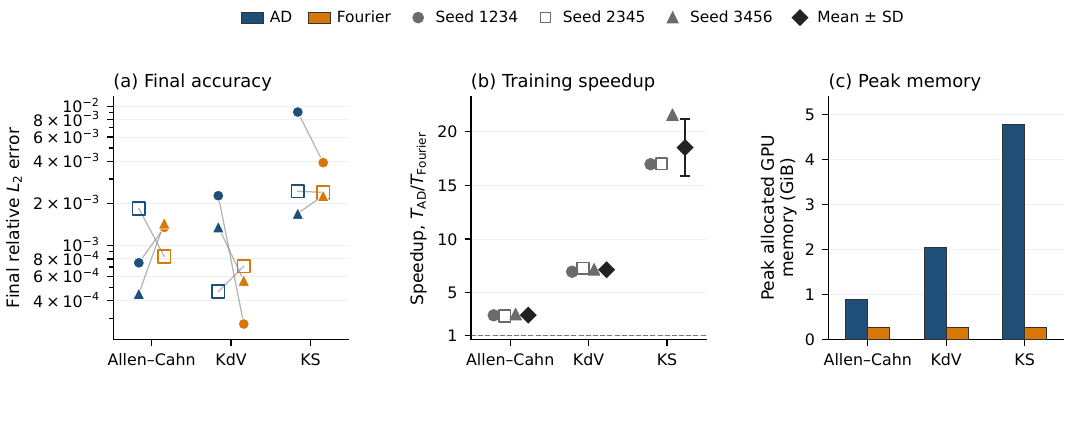}
    \caption{Quantitative comparison for the Causal PINN experiments.
    (a) Final-model relative \(L_2\) errors for the three paired random
    seeds, with AD and Fourier results for the same seed connected within
    each equation. (b) Paired training speedup
    \(T_{\mathrm{AD}}/T_{\mathrm{Fourier}}\); individual markers denote
    the three seeds, while the diamond marker and error bar show the mean
    and sample standard deviation. The dashed line denotes a speedup of
    one. (c) Peak allocated GPU memory for AD and Fourier spectral
    differentiation; the peak-memory values are identical across the three
    seeds for each equation--method combination.}
    \label{fig:causal_summary}
\end{figure}

Figure~\ref{fig:causal_summary}(a) shows the paired final-model relative
\(L_2\) errors for the Causal PINN experiments. The final errors remain of
the same order for both differentiation procedures across Allen--Cahn, KdV,
and KS, while the paired ranking varies with the random seed. Fourier
differentiation gives the lower final error in five of the nine paired runs,
whereas AD gives the lower error in four. The aggregate values are
summarized in Table~\ref{tab:results_summary}; the paired experiments do
not show a consistent accuracy advantage for either spatial differentiation
procedure.

Figure~\ref{fig:causal_summary}(b,c) summarizes the computational results.
Fourier spectral differentiation reduces training time for all nine paired
Causal PINN runs, with mean speedups of
\(2.90\pm0.09\times\), \(7.13\pm0.16\times\), and
\(18.52\pm2.65\times\) for Allen--Cahn, KdV, and KS, respectively. Peak
allocated GPU memory for AD is approximately \(0.908\), \(2.059\), and
\(4.784\) GiB across the three equations, whereas the Fourier
implementation remains near \(0.285\) GiB. The corresponding memory
reductions are \(68.7\%\), \(86.2\%\), and \(94.1\%\).

\begin{figure}[!htbp]
    \centering
    \includegraphics[width=\linewidth]
    {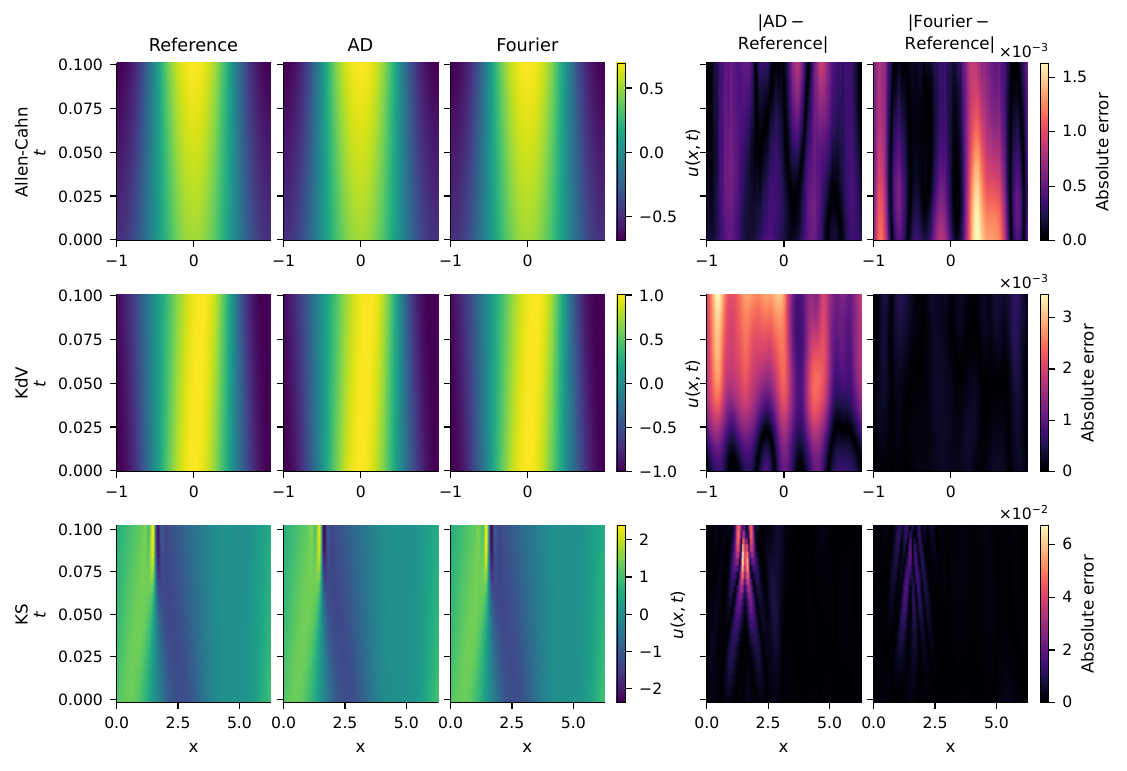}
    \caption{Reference, AD, and Fourier Causal-PINN solutions and
    pointwise absolute errors for seed 1234. Rows correspond to
    Allen--Cahn, KdV, and KS. For each equation, the three solution panels
    share one color scale and the two error panels share another.}
    \label{fig:causal_fields}
\end{figure}

Figure~\ref{fig:causal_fields} compares the reference solutions with the
final Causal PINN predictions for seed 1234. Both differentiation
procedures reproduce the principal space--time structures of the displayed
Allen--Cahn, KdV, and KS reference solutions.

%% file: sections/05_discussion.tex
\section{Discussion}
\label{sec:discussion}

\subsection{Computational Interpretation}

In the AD baseline, the required spatial derivatives are constructed through
repeated or nested differentiation of the network output, whereas the Fourier
implementation reuses one transformed representation across the spatial
derivative orders required by a given residual. The lower end-to-end training
time and peak allocated memory observed across both the standard and Causal
PINN experiments are consistent with this difference in derivative evaluation.

The cost of the spatial derivative path does not by itself determine the
end-to-end training speedup. The resulting gain also depends on how much of
the total training cost is attributable to derivative evaluation relative to
network evaluation and other operations. A related effect was reported for Fourier filter-based PhyCRNet: in its two-dimensional viscous Burgers and FitzHugh--Nagumo reaction--diffusion experiments, the time-series module dominated the training cost, limiting the overall computational benefit of the Fourier derivative calculation~\cite{xiao2022fourier}. The paired measurements in the present
study therefore characterize the effect of the spatial derivative backend at
the level of the complete training loop rather than the differentiation
operation in isolation.

Within the Causal PINN benchmarks, both the number of required spatial
derivative terms and the maximum derivative order increase from Allen--Cahn
to KdV and KS, together with larger paired speedups and higher peak memory
requirements for AD. Because the governing equations, nonlinear terms, and
derivative sets vary simultaneously across these benchmarks, the experiments
do not isolate the effect of derivative order or derivative count. The
observed sequence therefore constitutes a cross-equation empirical trend
rather than a controlled scaling law.

\subsection{Relation to Existing Derivative-Evaluation Methods}

Approaches that reduce or reformulate the use of spatial AD make different
choices in derivative discretization and neural representation. CAN-PINN couples AD with local numerical differentiation, while SK-PINN
evaluates derivatives through smoothing-kernel discretization
~\cite{chiu2022canpinn,pan2025skpinn}. SINN makes a broader
representational change by predicting spectral coefficients and performing
spatial differentiation directly in the spectral domain
~\cite{yu2026sinn}. Fourier filter-based PhyCRNet provides a direct precedent for Fourier differentiation of neural-network outputs: its predicted fields are transformed
to Fourier space, differentiated spectrally, and returned to physical space
before construction of the physics-informed loss~\cite{xiao2022fourier}.

The present experiments focus on a controlled computational comparison
within coordinate-based physical-space PINNs. Within each paired run, the
network initialization, neural representation, temporal differentiation,
temporal sampling schedule, optimizer settings, and training length are
matched, while only the spatial differentiation procedure is changed between
AD and Fourier spectral differentiation. The comparison is performed under
both the standard PINN and Causal PINN formulations and directly measures
end-to-end training time, peak allocated memory, and solution error. Across
all tested configurations, Fourier differentiation reduces end-to-end training
time and peak allocated memory, while final relative \(L_2\) accuracy remains
seed-dependent and shows no consistent advantage for either differentiation
procedure.

\subsection{Scope and Limitations}

The present comparison is restricted to one-dimensional PDEs on uniform
periodic spatial grids, where Fourier spectral differentiation can be applied
directly to the sampled network output. The experiments do not establish the
same computational trade-off for non-periodic domains, irregular spatial
discretizations, or higher-dimensional problems. The quantitative speedups
are also specific to the experimental configuration and hardware used here
and may change with network size, spatial resolution, implementation, and
hardware. The Causal PINN experiments use a fixed causality parameter
\(\epsilon=0.1\) and do not evaluate the causality-parameter annealing strategy
used by Wang et al.~\cite{wang2024causal}.

A controlled study that varies the spatial-derivative workload within a fixed
PDE setting could isolate the scaling behavior more directly. Extending the
paired comparison to higher-dimensional periodic PDEs would also test whether
the observed runtime and memory advantages persist as the spatial
discretization and Fourier transforms become larger.

%% file: sections/06_conclusion.tex
\section{Conclusion}
\label{sec:conclusion}

We performed a controlled comparison of spatial automatic differentiation
and Fourier spectral differentiation in physical-space PINNs, with all
components of each paired training setup held fixed except the spatial
differentiation procedure. The comparison was carried out in standard PINNs
for the Allen--Cahn and KdV equations and in Causal PINNs for Allen--Cahn,
KdV, and KS. Across these five equation--framework settings, Fourier spectral
differentiation yielded mean paired speedups ranging from \(2.90\times\) to
\(18.52\times\) and reduced peak allocated GPU memory by
\(68.7\%\)--\(94.1\%\). Final relative \(L_2\) errors remained of the same
order for the two differentiation procedures, with no consistent accuracy
advantage across the paired runs. For the one-dimensional periodic problems
considered here, changing the spatial differentiation procedure from AD to
Fourier spectral differentiation substantially reduced the computational
cost of training without a consistent loss of solution accuracy.